\documentclass{article}

\usepackage{arxiv}

\usepackage[utf8]{inputenc} 
\usepackage[T1]{fontenc}    
\usepackage{hyperref}       
\usepackage{url}            
\usepackage{booktabs}       
\usepackage{amsfonts,amsmath,amssymb}       
\usepackage{nicefrac}       
\usepackage{microtype}      
\usepackage{graphicx}

\usepackage{color, colortbl}

\usepackage{mathptmx,amsthm}
\usepackage{xcolor}
\usepackage{multirow}

\newcommand{\bdesc}{\begin{description}}
\newcommand{\edesc}{\end{description}}
\newcommand{\beqna}{\begin{eqnarray}}
\newcommand{\eeqna}{\end{eqnarray}}
\newcommand{\benum}{\begin{enumerate}}
\newcommand{\eenum}{\end{enumerate}}
\newcommand{\bitem}{\begin{itemize}}
\newcommand{\eitem}{\end{itemize}}
\newcommand{\bfig}{\begin{figure}}
\newcommand{\efig}{\end{figure}}
\newcommand{\cl}{\ensuremath{C_l}}
\newcommand{\cs}{\ensuremath{C_s}}
\newcommand{\cp}{\ensuremath{C_p}}
\newcommand{\egeom}{\ensuremath{E_{\text{geom}}}}
\newcommand{\optscale}{\ensuremath{s_{\text{opt}}}}

\newcommand{\scale}{\ensuremath{s}}
\newcommand{\rscale}{\ensuremath{r}}
\newcommand{\kscale}{\ensuremath{k}}
\newcommand{\mulscaleapproach}{\textit{Multiscale}}
\newcommand{\optscaleapproach}{\textit{Optimal-scale}}
\newcommand{\salmap}{\ensuremath{\{\cl,\cs,\cp\}}}
\newcommand{\centroid}{\ensuremath{\bar{G}}}

\newcommand{\bdist}{\ensuremath{d_{\text{BD-Img}}}}
\newcommand{\edist}{\ensuremath{d_{\text{EMD-Img}}}}

\newcommand{\ssim}{\ensuremath{\mathbf{s}_{\text{SSIM}}}}

\newcommand{\emd}{\ensuremath{d_{\text{EMD-Info}}}}

\newcommand{\eigsum}{\ensuremath{\mathbf{S}}}
\newcommand{\vt}{\ensuremath{T_{\text{3DVT}}}}
\newcommand{\vtget}{\ensuremath{T_{\text{3DVT-GET}}}}

\newcommand{\cov}{\ensuremath{T_{\text{3DCM}}}}

\newcommand{\sg}{\ensuremath{\mathcal{S}_{Gm}}}
\newcommand{\sas}{\ensuremath{\mathcal{S}_{AgSm}}}
\newcommand{\ptc}{\ensuremath{\mathcal{P}}}
\newcommand{\ptcq}{\ensuremath{\mathcal{Q}}}
\newcommand{\kld}{\ensuremath{\delta_{KL}}}
\newcommand{\fvd}{\ensuremath{\delta_{FV}}}
\newcommand{\hmd}{\ensuremath{L_1}}

\newcommand{\commenttxt}[1]{}
\newcommand{\mysubheading}[1]{\vspace{0.25cm}\noindent\textbf{#1}}

\definecolor{Gray085}{gray}{0.85}
\usepackage{colortbl}
\newcolumntype{a}{>{\columncolor{Gray085}[.9\tabcolsep][0.9\tabcolsep]}c}

\theoremstyle{definition}
\newtheorem{definition}{Definition}[section]

\title{Augmented Semantic Signatures of Airborne LiDAR Point Clouds for Comparison}
\author{
  Jaya Sreevalsan-Nair\thanks{\texttt{jnair@iiitb.ac.in}} \hspace{1cm} Pragyan Mohapatra\\
  Graphics-Visualization-Computing Lab,\\
  International Institute of Information Technology Bangalore, Karnataka 560100, India \\
  \texttt{http://www.iiitb.ac.in/gvcl} 
}

\begin{document}
\maketitle

\begin{abstract}
  LiDAR point clouds provide rich geometric information, which is particularly useful for the analysis of complex scenes of urban regions. Finding structural and semantic differences between two different three-dimensional point clouds, say, of the same region but acquired at different time instances is an important problem. Comparison of point clouds involves computationally expensive registration and segmentation. We are interested in capturing the relative differences in the geometric uncertainty and semantic content of the point cloud without the registration process. Hence, we propose an orientation-invariant geometric signature of the point cloud, which integrates its probabilistic geometric and semantic classifications. We study different properties of the geometric signature, which are image-based encoding of geometric uncertainty and semantic content. We explore different metrics to determine differences between these signatures, which in turn compare point clouds without performing point-to-point registration. Our results show that the differences in the signatures corroborate with the geometric and semantic differences of the point clouds.
\end{abstract}


\section{Introduction}\label{sec:introduction}
Topographic Light Detection and Ranging (LiDAR) technology provides three-dimensional (3D) point clouds, which are used for reconstructing regions. While there has been an active interest in investigating what analyses the point clouds can exclusively provide~\cite{rottensteiner2009status}, it has limitations in its data acquisition process, amongst others. Airborne LiDAR technology suffers from the known problem of reproducibility of logistics of acquisition of point clouds to study temporal changes~\cite{jalobeanu2014uncertainty}. Thus, the point clouds of the same region recorded in two different time instances may have significantly different point densities and poses/orientations, thus leading to misregistration errors~\cite{jalobeanu2014uncertainty}. Such differences lead to the strict requirement of a registration process of  point-to-point matching between point clouds for further analysis. There also exist other gaps in comparison of different point clouds. There are several methods to compare either their geometric properties, such as Fisher modified vector~\cite{ben20183dmfv}, or semantic compositions, such as Hamming distance of normalized distributions or symmetric Kullback Leibler (KL) divergence. Measures that integrate both geometric and semantic differences are rare.

\begin{figure}
    \centering
    \includegraphics[width=\columnwidth]{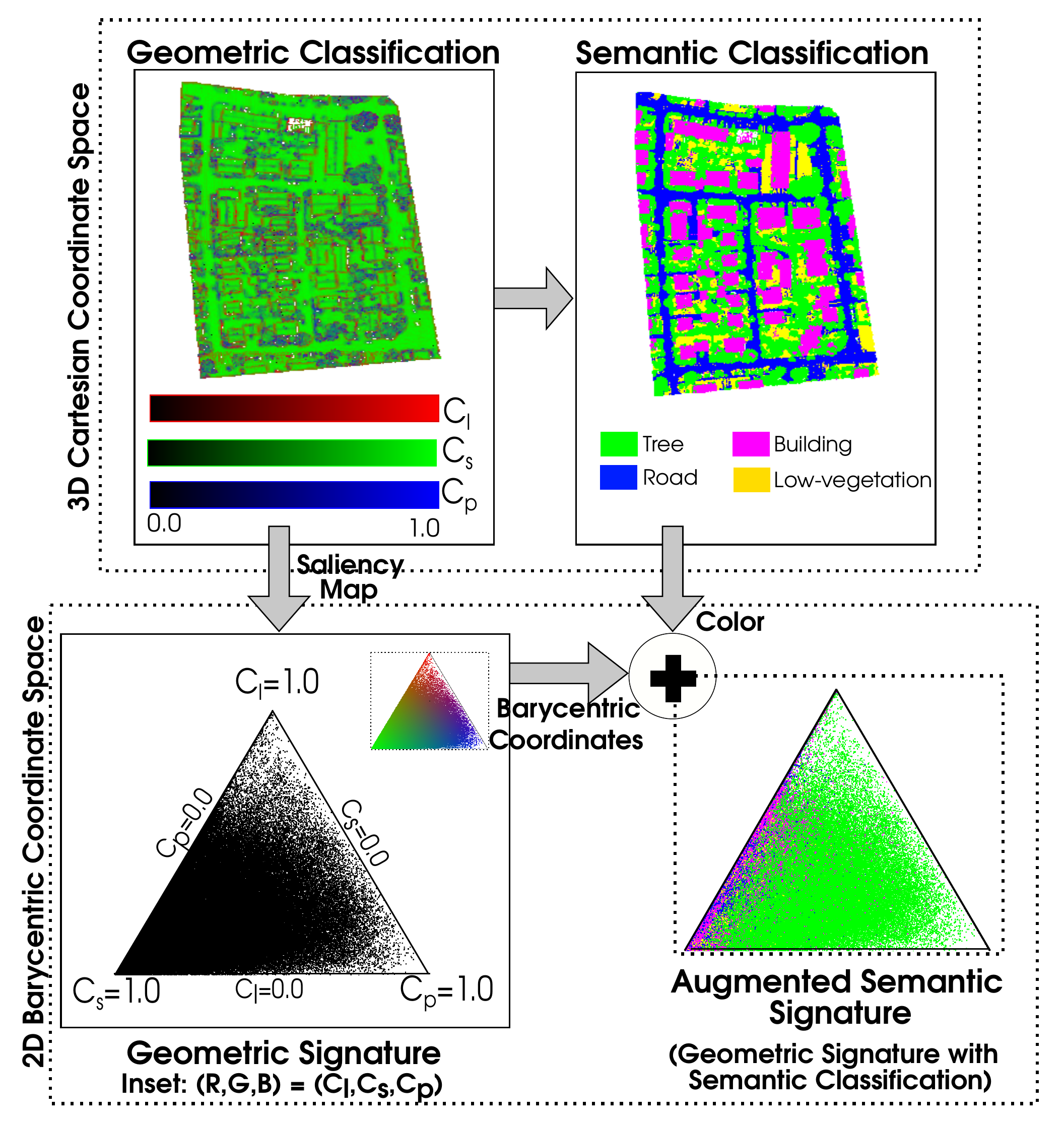}
    \caption{Overview of our proposed method of using probabilistic geometric classification for generating our proposed geometric signature of an airborne LiDAR point cloud. We generate the augmented semantic signature by adding semantic classification. The data is of Area-3 of the Vaihingen data~\cite{rottensteiner2012isprs}, the signatures are generated using $\cov$, with \mulscaleapproach~approach, and the visualizations are point renderings. (Top row) Top view of the 3D point cloud.}
    \label{fig:overview}
\end{figure}

To alleviate the need for registration, the point cloud could be projected to a different coordinate system by preserving the relationships between points, but not their absolute position coordinates. Hence, we propose the use of coordinate space transformations, which enable us to create a rotation-invariant global signature of the point cloud, as shown in the overview of our work (\S Figure~\ref{fig:overview}). This signature is the image of the two-dimensional (2D) barycentric representation of the entire point cloud, constructed using multiscale local geometric descriptors~\cite{sreevalsannair2017local}. We call it the \textit{geometric signature} $\sg$ of the point cloud, which encodes probabilistic geometric classification into \textit{geometric classes}, namely, line-, surface-, and (critical) point-type features. A variant of the signature incorporates semantic classification, which is referred to as the \textit{augmented semantic signature} $\sas$. It corresponds to a variant of the augmented semantic classification, which labels each point with a 2-tuple comprising of geometric and semantic class labels~\cite{kumari2015interactive}. Semantic classes are dataset-specific, and include buildings, trees, low-vegetation, and roads, in most urban datasets, e.g., Vaihingen site~\cite{rottensteiner2012isprs}. $\sg$ must satisfy the requirements of metric-invariance, orientation-invariance, and, additionally, class-dependence in the case of $\sas$. These requirements ensure that the signature contains the geometric classification of the point cloud, and optionally, its semantic classification necessary for its contextual understanding~\cite{weinmann2015semantic}.

One of the motivating applications of $\sg$ of the point clouds is their comparison. Comparing  $\sg$ of point clouds implies finding the differences between their geometric classifications, arising from the differences in data acquisition (e.g., change in sampling density, orientation, etc.), or processing (e.g., use of different local geometric descriptors (LGDs) for geometric classification, multiscale approach, etc.). We are interested in point cloud comparison, which integrate both geometric and semantic information to indicate an event, e.g., deforestation in urban regions over time. 

Apart from visualizing the change in signatures, we propose metrics to quantify their differences using approaches based on root mean square deviation, image processing, and information theory. We propose a new definition for Shannon entropy computed using probabilistic geometric classification to serve two purposes. Firstly, this measure referred to as \textit{saliency map-based Shannon entropy} is used in the multiscale approach for computing local geometric descriptors. Secondly, it is used to determine the information-theoretic difference between $\sg$ of point clouds. Overall, $\sg$ of a point cloud reveals the uncertainty in its geometric classification.

Our contributions are:
\benum
\item We propose a novel geometric signature $\sg$ of an airborne LiDAR point cloud using probabilistic geometric classification, which is a global feature descriptor.
\bitem \item Its variant, the augmented semantic signature, incorporates its semantic classification. \eitem
\item We propose a saliency map-based Shannon entropy for determining the optimal scale of local geometric descriptors and for point cloud comparison.
\item We propose quantitative metrics computed from the geometric signature for point cloud comparison.
\eenum

\section{Related Work}\label{sec:relatedwork}
In this section, we discuss the relevant literature on signatures of point clouds and on point cloud comparisons.

\mysubheading{Geometric Signatures of Point Clouds:} Histograms have been widely used as signatures of point clouds, e.g., persistent feature histograms~\cite{rusu2008aligning}, histograms of the distribution of geometric distances~\cite{mahmoudi2009three}. Spin image is a rotation-invariant signature of point clouds~\cite{velizhev2012implicit}, which is the distribution of surface normals in connected components. These have been used for shape recognition, classification, and reconstruction.

Similar to our method, a barycentric coordinate representation of pointwise local geometric descriptor has been used for the class-based analysis of point clouds in geomorphology~\cite{brodu20123d}. This signature, referred to as the \textit{dimensionality density diagram} (DDD) uses normalized eigenvalues, and collates the signatures across multiple scales in a feature vector for semantic classification. In our work, the multiscale integration occurs before computation of eigenvalue-based saliencies~\cite{sreevalsannair2017local}, used as barycentric coordinates, and the signatures are used for determining structural and semantic changes in point clouds.

\mysubheading{Aggregated Local Feature Descriptors:} Given that 3D LiDAR point clouds implicitly encode a surface, they are 2.5D data characterized by surface feature descriptors. 3D surface feature descriptors can be categorized into the global and the local feature descriptors. The local feature descriptors encode the characteristics of the local neighborhoods around specific keypoints. They are, therefore, more suitable for recognizing partially visible or incomplete objects in a cluttered scene with occlusions. In contrast, the global feature descriptors ignore shape details and require a priori segmentation of the object from the scene. Thus, they have difficulty recognizing partially visible or incomplete objects from cluttered scenes. Hence, global features, that aggregate local features, also embed the geometric information.

The local geometric descriptors used for LiDAR point clouds include covariance tensor~\cite{gross2006extraction}, $\cov$, and anisotropically diffused tensor $\vt$, obtained from tensor voting~\cite{sreevalsannair2017local}. Eigenvalue decomposition of the covariance tensor~\cite{hoppe1992surface} provides point-wise features used for tangent plane estimation~\cite{hoppe1992surface}, geometric reconstruction~\cite{gross2006extraction,sreevalsannair2018contour}, and semantic classification~\cite{weinmann2015semantic}. Kernel heat signatures~\cite{sun2009concise}, or local surfaces at each point (multiscale features~\cite{li2005multiscale}, kernel correlation~\cite{shen2018mining}) are other widely use local descriptors for general point clouds. Rotational Projection Statistics (RoPS)~\cite{guo2013rotational} and, Tri-Spin-Image (TriSI)~\cite{guo2015novel} are robust to disturbances such as, noise, varying point density, clutter, and occlusion. RoPS is a local surface descriptor computed from multiple viewpoints, and TriSI is computed with respect to three orthogonal axes. However, they require the triangulation of the unordered point clouds, which is an overhead in real applications. Binary shape context descriptor is a robust descriptor computed on a scattered point cloud that encodes both the projection point density and distances from the local surface~\cite{dong2017novel}.

Fisher vectors (FV)~\cite{perronnin2010improving} and Vector of Locally Aggregated Descriptors (VLAD)~\cite{jegou2010aggregating} are information-based aggregates used as global feature descriptors. FVs have been computed as global descriptors of images using the local descriptors and the Fisher kernel~\cite{jaakkola1999exploiting}. The FV is the normalized gradient vector of the local descriptors. A Gaussian Mixture Model (GMM) is used as the probability density function used for modeling the local descriptors. The gradient vector is first computed from the log-likelihood of the sample with respect to the model parameters  (i.e., weight, mean and, covariance). The gradient vector is normalized using the Cholesky factor of the Fisher information matrix~\cite{jaakkola1999exploiting}. The Fisher kernel of two images is a similarity measure computed as the dot product of FVs of the two images. The kernel could also be computed with a selected generative model~\cite{sanchez2013image}. The VLAD, which is a global feature descriptor of an image, is computed using a codebook of clusters of SIFT (scale-invariant feature transform) descriptors in images, and accumulation of residuals between descriptors and corresponding cluster center. The VLAD has been further extended to point clouds to a deep learning network for place recognition~\cite{angelina2018pointnetvlad}. Similar to FV used for point cloud analysis~\cite{ben20183dmfv} and VLAD, our proposed $\sg$ is aggregated from local geometric descriptors explicitly, but using coordinate space transformation. Our $\sg$ encodes shapes of local neighborhoods, which implies the information of the geometric classification. Thus, $\sg$ is used for comparisons, different from the uses of the state-of-the-art for object detection and localization.

\mysubheading{Point Cloud Comparison:} A common technique for performing change detection in a point cloud is to compute its distance to a 3D reference model. The reference model can either be theoretical or simulated from real data. Point-to-mesh and mesh-to-mesh distances have been very well studied and demonstrated by Metro~\cite{cignoni1998metro}, Mesh~\cite{aspert2002mesh}, and other tools. However, the learning curve for using these software applications is steep. Also, 3D shape reconstruction of a point cloud is computationally intensive and hence, is neither efficient nor scalable. An out-of-core method has been used for change detection in massive point clouds, without reducing the raw data~\cite{richter2013out}. This involves computing the distance of each point to its closest point, which quantifies the change in geometry, using a memory-efficient multi-data octree. But, constructing such an octree for every comparison is inefficient.

Direct comparison of point clouds can be made with one-to-one point mapping by using robust distance measures, such as the Gromov-Hausdorff (GH) distance~\cite{memoli2004comparing}. GH distance generalizes the Hausdorff distance between two compact metric spaces, including probability measures and all isometric embeddings. It can be used for comparing LiDAR point clouds, which may have rotational transformations in follow-up data acquisitions, but after isometric transformations. However, computing a discrete approximation of GH distances using pair-wise geodesic distances is inefficient in large-scale point clouds, such as in airborne LiDAR. Recently, a feature set derived from local geometric descriptors, namely the covariance tensor, has been used for change detection in airborne LiDAR point clouds~\cite{tran2018integrated}. The feature set used in random forest (RF) classifier stores the information of the change in each point between two epochs and, thus, is further used to classify a specific pre-determined change in buildings or trees in the dataset. This method, however, requires registration between epochs. Our novel method quantifies integrated changes in semantic composition and geometric information, as in these methods.

\mysubheading{Multiscale Approaches:} There are two classes of approaches of integrating multiple scales in LiDAR point cloud processing, namely, determining an optimal scale and aggregating specific features computed from multiple scales. Generally, scales are chosen from a specific range, which can be fixed or adaptive across points. The use of \textit{optimal} scale in semantic classification enables the use of \textit{adaptive} scales. The optimal scale is where Shannon entropy of geometric features in the local neighborhood is minimum for the point~\cite{demantke2011dimensionality}. The feature vector for semantic classification has been widely generated at the optimal scale for supervised learning, e.g., using random forest classifier~\cite{weinmann2015semantic}. Aggregating specific features in a fixed range of scales is done by averaging across scales for saliency maps for geometric classification or reconstruction~\cite{keller2011extracting,sreevalsannair2017using,sreevalsannair2017local}. Here, we experiment with aggregation as well as optimal scale method for computing local descriptors needed for $\sg$.

Other recent multiscale methods include a scale-invariant method for generating hierarchical point clusters by aggregating scales in a multiscale and hierarchical point classification method~\cite{wang2014multiscale}. Latent Dirichlet Allocation (LDA) and AdaBoost classifiers are then used for feature extraction from the clusters and classification, respectively. A variant of this method uses natural exponential function thresholds for point-cluster extraction, and joint LDA and sparse coding (SCLDA) for feature extraction~\cite{zhang2016multilevel} for airborne LiDAR point clouds of complex scenes. To add contextual information, e.g., topology, in classification results,  an RF classifier has been integrated into a Conditional Random Field (CRF) framework for urban objects classification~\cite{niemeyer2014contextual}. Multiscale features accumulated in the CRF as a long feature vector have been used in combination with RF classifier~\cite{hackel2016fast} and Markov RFs~\cite{thomas2018semantic}. 

\mysubheading{Barycentric Coordinate Representation:} Barycentric coordinate system has been widely used in point cloud processing, e.g., for point-in-triangle test in various applications~\cite{battke1997fast}, and for curvature estimation~\cite{digne2011scale}. Barycentric coordinate representation has been used in triangle-occlusion tests for texture mapping of triangular meshes of LiDAR point clouds~\cite{zalama2011effective}. Generalized barycentric coordinate representation has been used in geometric modeling, for instance, in applications of interpolation and deformation~\cite{langer2008generalized}. 

\section{Methodology}\label{sec:method}
Our objective is to generate a signature of the entire point cloud, which is invariant to the orientation of the point cloud, sampling density, and the metric system used for data capture. The motivating application is to use our proposed geometric signature $\sg$ for finding the differences in geometric uncertainty and semantic content of two different LiDAR point clouds. The rationale for finding new measures of difference between point clouds is to find an integrated one, which captures the changes in point clouds holistically.

We propose the use of probabilistic geometric classification~\cite{sreevalsannair2017local} to project the point cloud to the 2D barycentric coordinate space. The barycentric coordinate space, thus, shows the information of geometric classification and is referred to as \textit{geometric signature} $\sg$. Upon adding the semantic classification information to the geometric signature, we create its variant, called as \textit{augmented semantic signature} $\sas$. 

\theoremstyle{definition}
\begin{definition}{} A \textit{geometric signature} $\sg$ of a point cloud $\ptc$ is an image of the points projected in the barycentric coordinate space, using the probabilistic geometric classification.
\end{definition}
\begin{definition}{} An \textit{augmented semantic signature} $\sas$ of a point cloud $\ptc$ is its geometric signature which is colored as per its semantic classification.
\end{definition}

$\sg$ represents the probabilistic geometric classification, and not the widely used deterministic geometric classification~\cite{kumari2015interactive}. Thus, our proposed geometric signature of a point cloud is a global feature descriptor, which aggregates local geometric descriptors. The point rendering in the barycentric coordinate space gives us the scope to use visual encodings such as color, size, and shape of the point to display its attributes\footnote{However, the use of visual encodings of size and shape of the point (glyph) is subject to the constraint of permissible visual clutter in the image.}, similar to that in the 3D Cartesian coordinate space. When the color of the point corresponds to its semantic class, we \textit{augment} the geometric signature with the semantic classification, thus creating $\sas$,  which corresponds to augmented semantic classification~\cite{kumari2015interactive}. Unlike the existing signatures, $\sg$ does not require intermediate 3D representation, point registration between point clouds, or point-wise comparisons. This is because we measure only the overall differences between the point clouds in geometric and semantic classifications. Even though $\sas$ uses label information, our method does not explicitly perform semantic classification, unlike similar methods~\cite{tran2018integrated}.  Our work is novel in using 2D image comparisons, which is traditionally used in 2D imagery in remote sensing~\cite{tran2018integrated}, for 3D data.

\subsection{Probabilistic Geometric Classification}\label{sec:method_probgeomclass}
The local geometric descriptors are predominantly used as features for semantic classification, which internally uses geometric classification~\cite{kumari2015interactive,keller2011extracting}. The geometric classification of a point is based on the cylindrical, disc, or spherical shape of points in the local neighborhood of the point. The shape of the neighborhood thus gives line-, surface- or (critical) point-type features, respectively. The shape is determined using the eigenvalue analysis of the local geometric descriptors. The saliency map is a 3-tuple that gives the likelihood of a point belonging to each of the three geometric classes, namely, line-, surface-, and point-type features. The saliency map is $\salmap(x)$, such that $\cl(x)+\cs(x)+\cp(x)=1.0$. Thus, probabilistic geometric classification is represented by the saliency map. Geometric classification is inherently probabilistic~\cite{sreevalsannair2017local}, but it can be rounded off to make it deterministic~\cite{kumari2015interactive} by using the maximum saliency values. Augmented semantic classification involves tuple of labels from semantic and deterministic geometric classifications~\cite{kumari2015interactive}. Here, we use augmented semantic classification as a tuple of the semantic label and the saliency map of the probabilistic geometric classification.

Augmented semantic classification is a terminology introduced by Kumari and Sreevalsan-Nair~\cite{kumari2015interactive}, where each point in the point cloud has a tuple of labels, namely the semantic and geometric classes. The uniqueness of the combination of semantic and geometric classes in the tuples ensures clear separability of augmented semantic classes. For instance, augmented semantic classes include ``line-type feature on a building,'' ``point-type feature on a tree,'' etc. Thus, the augmented semantic classification is of finer-granularity compared to its constituent classifications.

Geometric classification of LiDAR point clouds is determined by the local geometric descriptors~\cite{sreevalsannair2017local,sreevalsannair2018contour}. Other than covariance tensor, tensor voting based on the perceptual organization using Gestalt principles~\cite{medioni2000tensor} also is used to construct a positive semidefinite second-order tensor for unoriented points~\cite{park2012multi}. Such a tensor is not the same as the covariance tensor, as the former is generated in normal space and the latter, in tangent space~\cite{sreevalsannair2017local}. Hence, anisotropic diffusion is applied to get the geometrical features similar to those of covariance tensor~\cite{sreevalsannair2017local,wang2013anisotropic}. After applying anisotropic diffusion, $\vt$ is further modified using gradient energy tensor obtained from 2D tensors in cylindrical neighborhood~\cite{sreevalsannair2018contour,sreevalsannair2017using}. This tensor, referred to as $\vtget$, has a higher percentage of point-type features, especially for the tree class.

The local neighborhood used for descriptor computation is conventionally determined either as \textit{nearest k neighbors} for $\kscale\in\mathbb{Z}^+$, or as a set of points within the \textit{spherical neighborhood} of radius $\scale\in\mathbb{R}^+$. In each case, the size of the neighborhood, $\kscale$ or $\scale$, is considered as the \textit{scale}. It has been widely understood that the outdoor environmental scenes captured by airborne LiDAR point clouds have larger uncertainties involved in scene understanding. Hence, a single scale analysis of the local neighborhood of the points is insufficient~\cite{sreevalsannair2017local,hackel2016fast,thomas2018semantic}. Therefore, multiscale approaches are used to aggregate the information across scales. In this work, we use scales from both $k$ nearest and spherical neighborhoods. We then integrate the scales in each case using two different approaches, namely, the optimal and aggregated methods. An \textit{optimal} scale can be computed using minimum Shannon entropy values across different scales~\cite{weinmann2015semantic,demantke2011dimensionality}. An \textit{aggregated} method can be used by averaging saliency maps across different scales.

\subsection{Geometric Signatures}\label{sec:method_signature}
As the saliency map at a point is a set of probabilities that sum up to one, it is equivalent to barycentric coordinate representation of a point in the reference triangle\footnote{The reference triangle in 2D barycentric coordinate space is equivalent to the reference axes in the 3D cartesian coordinate system.} in the 2D barycentric coordinate space (\S Appendix~\ref{sec:bg-bcs}). Thus, our proposed $\sg$ maps points in 3D cartesian to 2D barycentric coordinate spaces. $\sg$ is influenced by factors such as choice of LGDs, choice of multiscale approaches, and the characteristics of the function/map between two coordinate spaces. $\sas$ is influenced by the semantic composition.

\begin{figure}
\centering  
\fbox{\includegraphics[width=0.9\columnwidth]{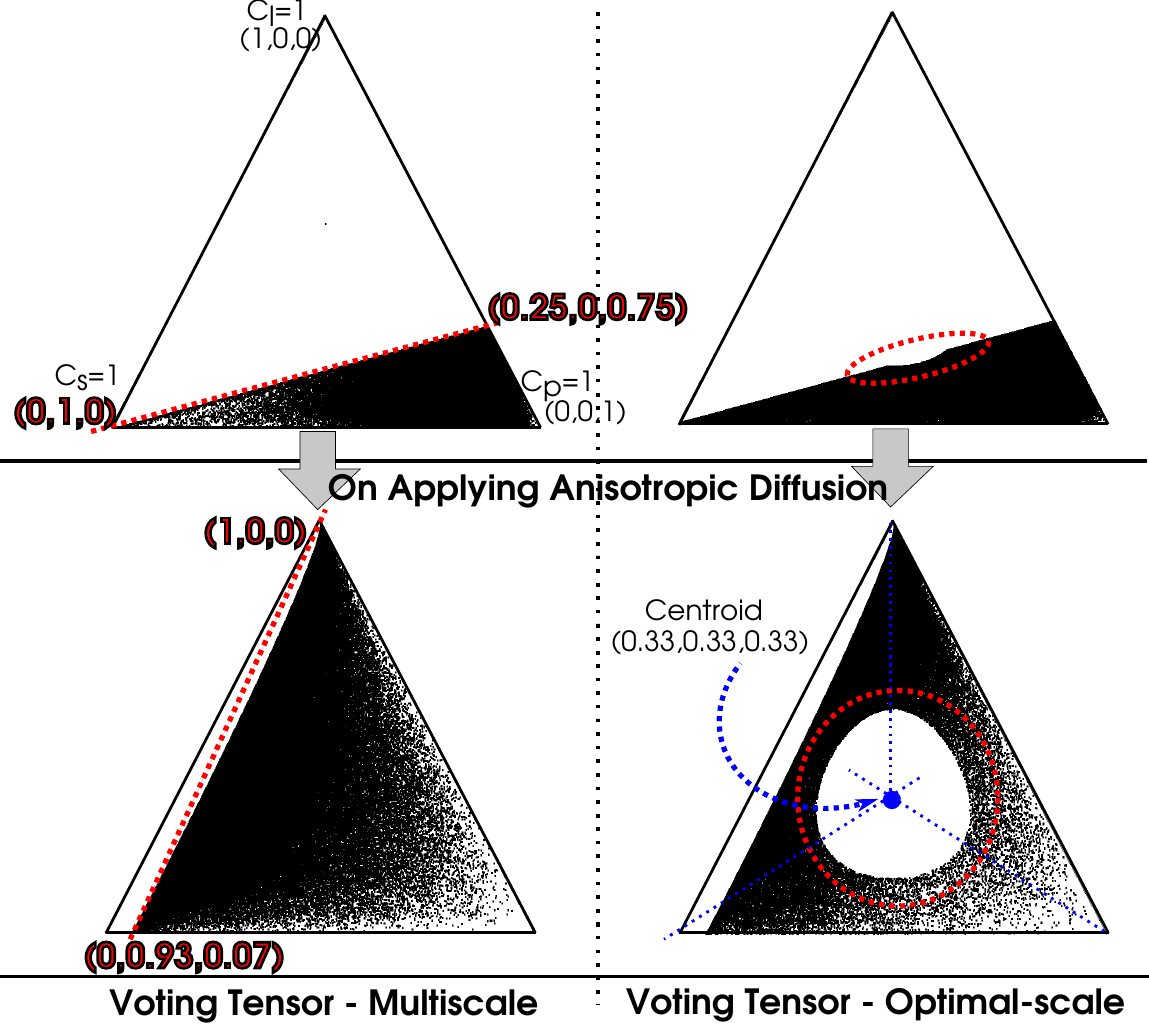}}
\caption{\label{fig:a2aniso} Patterns observed in the geometric signatures in tensor voting and optimal scale are shown in red dotted lines and ellipses, respectively, in the reference triangle in barycentric coordinate space. (Top left) The partitioning line between (0,1,0) and (0.25,0,0.75) occurs due to the computations of  tensor voting as ball tensors. (Bottom left) This line transforms to a curve between (1,0,0) and (0,0.93,0.07) upon applying anisotropic diffusion to get $\vt$. Upon using optimal scale in $\vt$, additionally (top right) a ``dent'' transforms to (bottom right) a ``hole'' in $\vt$ after applying anisotropic diffusion. The points located away from the centroid ($\frac{1}{3},\frac{1}{3},\frac{1}{3}$) and closer to the vertices have lower Shannon entropy, used for optimal scale determination. These images are generated for LiDAR point cloud of Area-3 of Vaihingen dataset~\cite{rottensteiner2012isprs} (323,896 points) for $\vtget$, a variant of $\vt$ after integrating GET, with same patterns in $\sg$ seen for $\vt$.}
\end{figure}

\subsubsection{Choice of Local Geometric Descriptors}
Covariance tensor $\cov$ and anisotropically diffused tensor using tensor voting $\vt$~\cite{sreevalsannair2017local} are LGDs of our interest. In the latter case, we use the tensor integrated with 2D gradient energy tensor~\cite{sreevalsannair2017using}, $\vtget$\footnote{We have used the computations for $\cov$, $\vt$, and anisotropic diffusion as given by Sreevalsan-Nair and Kumari~\cite{sreevalsannair2017using}, and for $\vtget$ as given by Sreevalsan-Nair et al.~\cite{sreevalsannair2018contour}.}. The choice of LGD gives characteristic geometric signatures, as the distribution of saliency maps in the LiDAR point cloud varies across different LGDs, e.g., $\cov$ gives more point-type features than $\vt$~\cite{sreevalsannair2018contour}. $\vt$ is computed using ball tensors, as initialization for unoriented points. Hence, $\vt$ is characterized by $(\cl\leq\cp+\cs)$ in $\vt$, since the ball tensor is computed in the normal space and not in the tangent space, as it is for $\cov$~\cite{sreevalsannair2017local}. The computation in normal space causes the vote from each neighbor to be a curvel or a plate tensor where the minor eigenvalue is 0.0~\cite{tong2004first}. Thus, when these votes add up, a constraint on $\cl$ builds up, which manifests as $(\cl\leq\cp+\cs)$. Using Equation~\ref{eqn:saliency} (\S Appendix~\ref{sec:bg-lgd}), we find the upper bound, $\frac{\cs}{\cl}\leq 3.0$, manifesting as the \textit{partitioning line} in the barycentric space. The partitioning line is between $(0,1,0)$ and $(0.25,0,0.75)$ (\S red dotted line in Figure~\ref{fig:a2aniso}). Anisotropic diffusion is performed on tensor voting to strengthen weak line-type features~\cite{wang2013anisotropic} and to make the tensor voting-based LGD substitutable for $\cov$~\cite{sreevalsannair2017local}. Anisotropic diffusion transforms the partitioning line in the signature of $\vt$ to the curved boundary between $(1,0,0)$ and $(0,0.93,0.07)$ (\S Figure~\ref{fig:a2aniso}). This pattern in $\sg$ demonstrates two known characteristics of $\vt$~\cite{sreevalsannair2017local,sreevalsannair2018contour}, which are that applying anisotropic diffusion increases the number of line-type features and reduces the number of point-type features in $\vt$. Thus, the $\sg$ gives visual validation of the characteristics of the tensor voting based LGD, $\vt$, especially the influence of anisotropic diffusion on it.

\subsubsection{Multiscale Approaches \& Novel Entropy Metric}
Here, we use both  multiscale aggregation of saliency maps as well as optimal scale determination to integrate multiple scales from a fixed range of scales. The multiscale aggregation refers to averaging of the saliency maps calculated at a range of scales~\cite{sreevalsannair2017local,sreevalsannair2017using} (\S Equation~\ref{eqn:saliency} in Appendix~\ref{sec:bg-lgd} gives saliency map computation for a single scale). Computing an optimal scale at each point is done by finding the local minimum of Shannon entropy computed of the local neighborhood of each point, across scales. Shannon entropy is computed based on either dimensionality~\cite{demantke2011dimensionality} or eigen-entropy~\cite{weinmann2014semantic}. The dimensionality method uses linearity, planarity, and scattering of $\cov$~\cite{demantke2011dimensionality}, whereas, the eigen-entropy method computes the same from normalized eigenvalues of $\cov$~\cite{weinmann2014semantic}. Thus, eigenvalue decomposition (EVD) of LGD is significant in multiscale approaches. In fact, saliency maps capture information of the EVD. Since the saliency maps at each point obey the partition to unity rule and is a variant of normalized eigenvalues, it can be used computing Shannon entropy\footnote{For interested readers, Weinmann et al. have explained in detail the use of eigenvalue-based features for use in Shannon entropy~\cite{weinmann2015semantic}.}. We propose a novel variant of Shannon entropy $\egeom$ using saliency maps, which gives uncertainty in probabilistic geometric classification. Our saliency map-based Shannon entropy at a point x is given as\footnote{As an operator, $\egeom=-\cl.ln(\cl)-\cs.ln(\cs)-\cp.ln(\cp)$.}, \\\centerline{$\egeom(x)=\sum\limits_{k\in\{l,s,p\}}-C_k(x).ln(C_k(x)).$}

\noindent Thus, we use the minimum value of $\egeom$ to select the optimal scale $\optscale$, and then use the saliency maps at $\optscale$. $\egeom$ is the measure of the uncertainty in probabilistic geometric classification, which we refer to as \textit{geometric uncertainty}.

We observe that the $\sg$ using optimal scale based on $\egeom$, have a distinct ``hole'' or empty region in the neighborhood of the centroid, where $\egeom$ is the maximum (\S Figure~\ref{fig:a2aniso}). This pattern occurs because the optimal scale is at local minimum of $\egeom$, corresponding to the saliency maps closer to the vertices of the reference triangle, i.e., where either $\cl$, $\cs$, or $\cp$ is 1. We observe the same pattern for $\cov$ (\S Section~\ref{sec:results}). In the case of optimal scale in tensor voting, we observe an additional hemispherical ``dent'' in the partitioning line in $\vt$, which transforms to the ``hole'', upon applying anisotropic diffusion (\S Figure~\ref{fig:a2aniso}). Thus, the pattern in the $\sg$ with $\vt$ shows the combined effects of using the optimal scale and anisotropic diffusion.

\subsubsection{Influence of Semantic Class in Signatures}
The semantic classification has conventionally been implemented using feature vectors derived from LGDs. If these features enable class separability in different classification methods, we can hypothesize that the different semantic classes, such as trees, low-vegetation, buildings, etc. seen in airborne LiDAR point clouds, in turn, have characteristic $\sg$. Our hypothesis stems from the observed differences in the composition of line-, surface-, and point-type features in each class~\cite{demantke2011dimensionality}. For instance, the tree and low- vegetation classes consist of predominantly point-type features~\cite{demantke2011dimensionality,sreevalsannair2017using}. The building class consists of a relatively higher proportion of surface-type features along with a lower proportion of line-type features~\cite{demantke2011dimensionality}, which is the characteristic of the roof surface and contour~\cite{sreevalsannair2018contour}. Our hypothesis is the basis of our motivating application of using $\sg$ for determining differences between two point clouds, including their semantic composition.

\subsubsection{Non-injective Surjective Function Map}
The probabilistic geometric classification is a many-to-one mapping, which implies that several points in the point cloud could have the same saliency map. This, in turn, causes those points to have the same barycentric coordinate representation. At the same time, \textit{all} points in the point cloud will have a saliency map, and hence a barycentric coordinate representation. Thus, the mapping for generating geometric signatures is a \textit{non-injective surjective function}. One of the implications of the non-injective function is that several points in the point cloud overlap in the $\sg$. This nature of the function has implications on the applicability of the signatures. A point to note with respect to the non-injective mapping in the $\sg$ is that, though the signature encodes spatial locality, it \textit{does not} linearly transform spatial locality. Thus, our proposed signature \textit{loses} the spatial context and hence, can be used \textit{only} for applications or analytical tasks to visualize the composition of the scene using the geometric properties of different semantic classes whose objects are present in the scene.

The sampling density of the point cloud influences the overlap of points in the signature. Thus, sparser point clouds tend to give \textit{even sparser} $\sg$. However, we can say that our $\sg$ are \textit{resilient} to sampling density, i.e., small differences in sampling density for a point cloud do not manifest as \textit{perceptual differences} in the signature (more in Section~\ref{sec:results}). 

\subsection{Comparing LiDAR Point Clouds}\label{sec:method_compare}
We visualize the $\sg$ of two different LiDAR point clouds to compare them qualitatively. For quantitative comparison, conventional methods use injective mapping after performing registration. Injective mapping refers to the one-to-one mapping of points between two point clouds, where the more accurate distances, such as, point-wise root mean square error (or difference) (RMSE), can be computed. Distribution-based metrics are apt for non-injective mapping, such as in the case of absence of registration. For determining differences in semantic composition, the difference between count distribution across different classes can be computed using distance metrics, such as, Hamming distance of normalized distributions and symmetric Kullback Leibler (KL) distance. In the same vein, we propose to use distributions of $\egeom$ in point clouds to quantify differences in geometric uncertainty. For integrating semantic composition and geometric uncertainty, we propose image-based metrics for both grayscale and colored images, for $\sg$ and $\sas$, respectively. 

\subsubsection{Difference in Uncertainty in Probabilistic Geometric Classification}
The distribution of $\egeom$ of the point cloud reflects the uncertainty in its probablistic geometric distribution. The plots of the probability distribution function (pdfs) of $\egeom$ of a point cloud give the uncertainties in geometric classification across different semantic classes when using different local geometric descriptors. Highest uncertainty in probabilistic geometric classification occurs at the $\centroid$ of the reference triangle of the barycentric coordinate space in $\sg$. When uncertainty is high, $\egeom$ is high, and points tend to be closer to $\centroid$. We compare two point clouds $\ptc$ and $\ptcq$ using the Earth Mover's Distance (EMD) \cite{qianwen2014comparison} between the pdfs of $\egeom$ of each point cloud. EMD determines the minimum cost of transforming a histogram into another, where higher EMD implies higher transformation cost. Thus, our information-theoretic distance measure is given as: \\\centerline{$\emd=d_{\text{EMD}}(\text{pdf}(\egeom(\ptc)),\text{pdf}(\egeom(\ptcq)))$.}

\subsubsection{Image-based Metric}
The signatures generated in \textit{congruent reference} triangles can be compared using the difference in the point distribution in the interior of the triangles. We store the \textit{graphical rendering} of the point cloud in the interior of the triangle as a pixellated image. Thus, the difference between $\sg$ of two point clouds gives the difference between geometric uncertainties, and between $\sas$ of the point clouds gives the difference in semantic composition. 

\mysubheading{For grayscale images of $\sg$:} Structural Similarity Index (SSIM) is used to measure the similarity between geometric signatures. SSIM measures the similarity score $\ssim\in[-1,1]$, rather than differences. A score of $+1$ indicates no difference between the two images. SSIM is conventionally computed using the luma of the image, i.e., the achromatic portion. Thus, we use SSIM to compute the similarity between the geometric uncertainty of two point clouds.

\mysubheading{For color images of $\sas$:} The distance between the discretized histograms of the probability density functions (pdfs) of the color channels give the difference in semantic classification in the point clouds. There are two groups of distance measures for a histogram, namely, bin-to-bin and cross-bin. The bin-to-bin computes the difference in the content of the bin with the corresponding bin of the second histogram, while the latter additionally considers the neighborhood values of the concerned bin. Conventionally used Bhattacharyya distance~\cite{naik2009scale} $\bdist$ is a bin-to-bin method which gives values  in the range $[0,1]$, where 0 and 1 mean a perfect match and maximum difference between the images, respectively. In addition to the symmetric variant of Bhattacharyya distance, we determine EMD $\edist$ between the images, which is a cross-bin distance measure. We use the normalized color histogram to determine the color distribution of the interior of the reference triangle in the image. Using a mask with size same as that of the triangle exclusively ignores the image background, thus reducing the discrepancies in the distances. These image metrics are sensitive to point sampling density, just as in the case of perceptual differences in the visualization of $\sg$ or $\sas$.

\section{Experiments}\label{sec:expt}
We have used the airborne LiDAR dataset for Area-2 and Area-3 in the Vaihingen site, from the ISPRS benchmark~\cite{rottensteiner2012isprs}. We have used the labeled dataset, with Area-2 (N = 266,675 points) having high-rise buildings (B = 69,097 points) with trees (T = 107,485 points), and Area-3 (N = 323,896 points) having independent small houses (B = 69,563 points), and trees (T = 135,499 points). As is done with all classification studies, we use the  height values of all points above ground, thus assuming a flat terrain. We compare Area-2 and Area-3 by using $\sg$ to capture the integrated differences in geometry and semantic composition.

\mysubheading{Choice of Multiscale Approach and Local Neighborhoods:} Our experiments include geometric signatures computed using the covariance tensor, $\cov$, and $\vtget$. $\vtget$ is computed using tensor voting, and modified using anisotropic diffusion and 2D gradient energy tensor of $\vt$~\cite{sreevalsannair2017local}. The LGDs have been computed using the following combination of local neighborhood types and multiscale approaches\footnote{The rationale behind choosing the specific combinations for our chosen LGDs is to compare the influence of the choice of the descriptor for computing the signatures, where the descriptor is already studied in the state-of-the-art.}: (a) k-nearest neighbors for optimal scale computation~\cite{weinmann2015semantic}, and (b) spherical neighborhood for averaging saliency maps over multiple scales~\cite{sreevalsannair2017local}. Hereafter, we refer to our chosen combination of multiscale approaches and local neighborhoods, as:

(a) ``\textbf{\mulscaleapproach}''~\textit{for saliency map in spherical neighborhoods average across multiple scales}, and 

(b) ``\textbf{\optscaleapproach}''~\textit{for saliency map from the local geometric descriptor at optimal scale determined specifically using $\egeom$\footnote{Eigen-entropy~\cite{weinmann2015semantic} and $\egeom$ give comparable results, as both use normalized eigenvalues.}, at multiple scales of k-nearest neighborhoods}.

For (a), the scales we have used are $\kscale_{min}=10$, $\kscale_{max}=100$, with step size\footnote{This is different from the step size used in Weinmann et al.~\cite{weinmann2015semantic}, $\Delta\kscale=1$, for semantic classification. Even with larger step size, $\Delta\kscale=10$, we have gotten good results for geometric classification for the same $\kscale$ range as for semantic classification. We get the benefit of a lesser number of computations here.} of $\Delta\kscale=10$. For (b), in the case\footnote{We have fitted the 3D (volumetric) data in a (canonical) bounding box of $[-1,1]\times[-1,1]\times[-1,1]$ using uniform scaling transformation, as done in computer graphics. In this normalized point cloud, we have used scales of $\rscale_{min}=0.009$, $\rscale_{max}=0.011$, with $\Delta\rscale=0.001$. Though we have methodically taken the same range of scales in this transformed space for all datasets, the scales in metric units vary from dataset to dataset. Hence, the different values for Area-2 and Area-3.} of Area-2, we have used the scales $\rscale_{min}=1.89m$, $\rscale_{max}=2.31m$, with step size of $\Delta\rscale=0.21m$, and in the case of Area-3, we have used the scales $rscale_{min}=1.92m$ , $\rscale_{max}=2.34m$, and $\Delta\rscale=0.21m$

\begin{figure}[t]
\centering
\fbox{\includegraphics[width=\columnwidth]{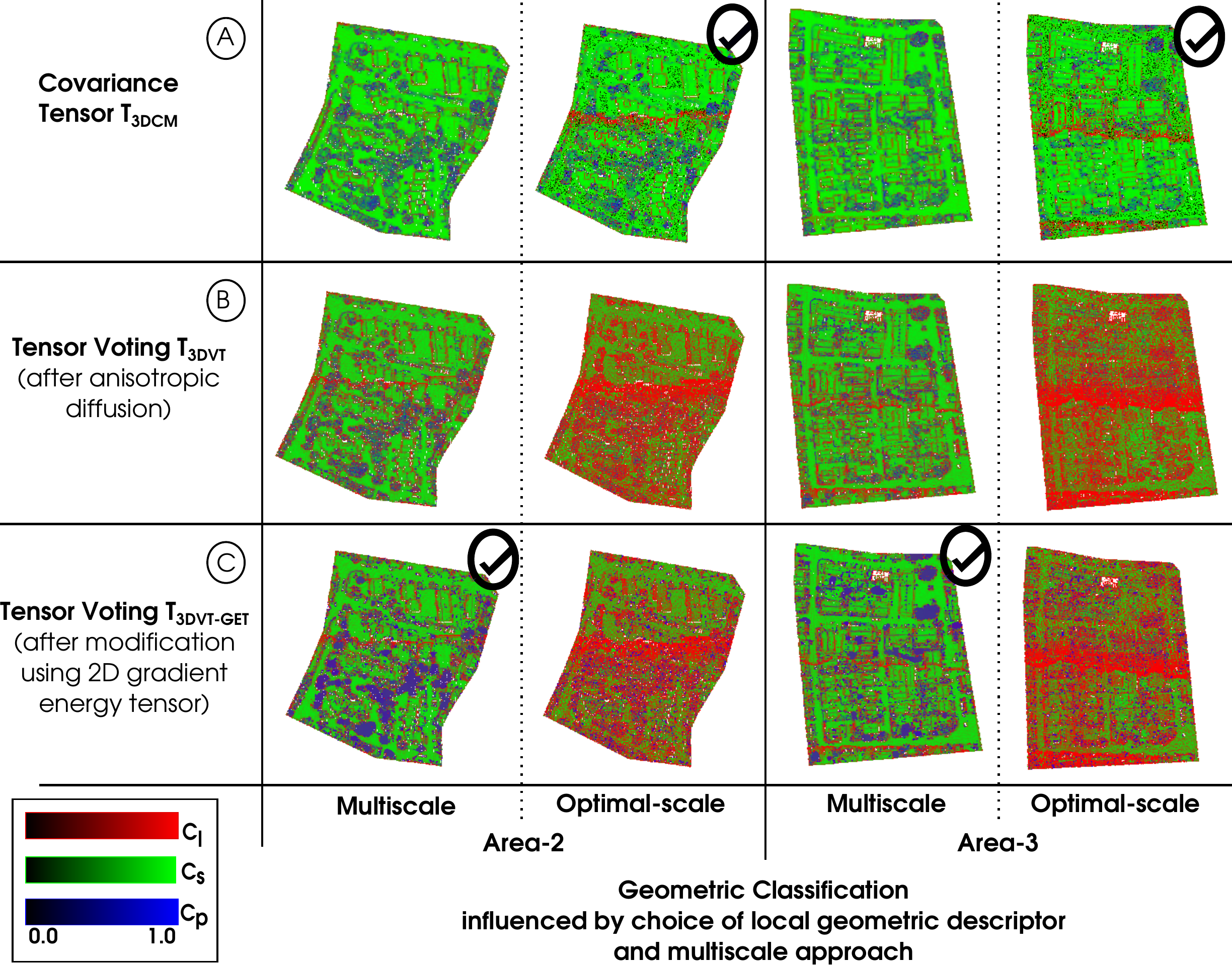}}
\caption{\label{fig:a2a3}
Top-down views of point rendering of Area-2 and Area-3 demonstrate the variations in geometric classification for different local geometric descriptors with different multiscale approaches. Each point is colored with saturation in red, green, and blue channels corresponding to $\cl$, $\cs$, and $\cp$ of its saliency map, respectively. $\cov$ and $\vtget$ are the local geometric descriptors used for generating signatures in our experiments, owing to anomalies in \optscaleapproach~$\vt$. The checks indicate the specific tensors used for further discussion of our results, owing to similarity in saliency maps.}
\end{figure}

\mysubheading{Choice of Local Geometric Descriptors:} The outcomes of probabilistic geometric classification of the descriptor tensors, $\cov$, $\vt$, and $\vtget$, in combination with each of the multiscale approaches, namely, the \mulscaleapproach~and the \optscaleapproach~approaches show anomalies in the case of \optscaleapproach~$\vt$ (\S Figure~\ref{fig:a2a3}). Hence, we use $\cov$ and $\vtget$ in our experiments. Since \mulscaleapproach~$\cov$, \optscaleapproach~$\cov$, and \mulscaleapproach~$\vtget$ show similar geometric classification results (\S Figure~\ref{fig:a2a3}), we use these tensors for further discussion in Section~\ref{sec:results}. 

\begin{figure}[htbp]
\centering  
\fbox{\includegraphics[width=\columnwidth]{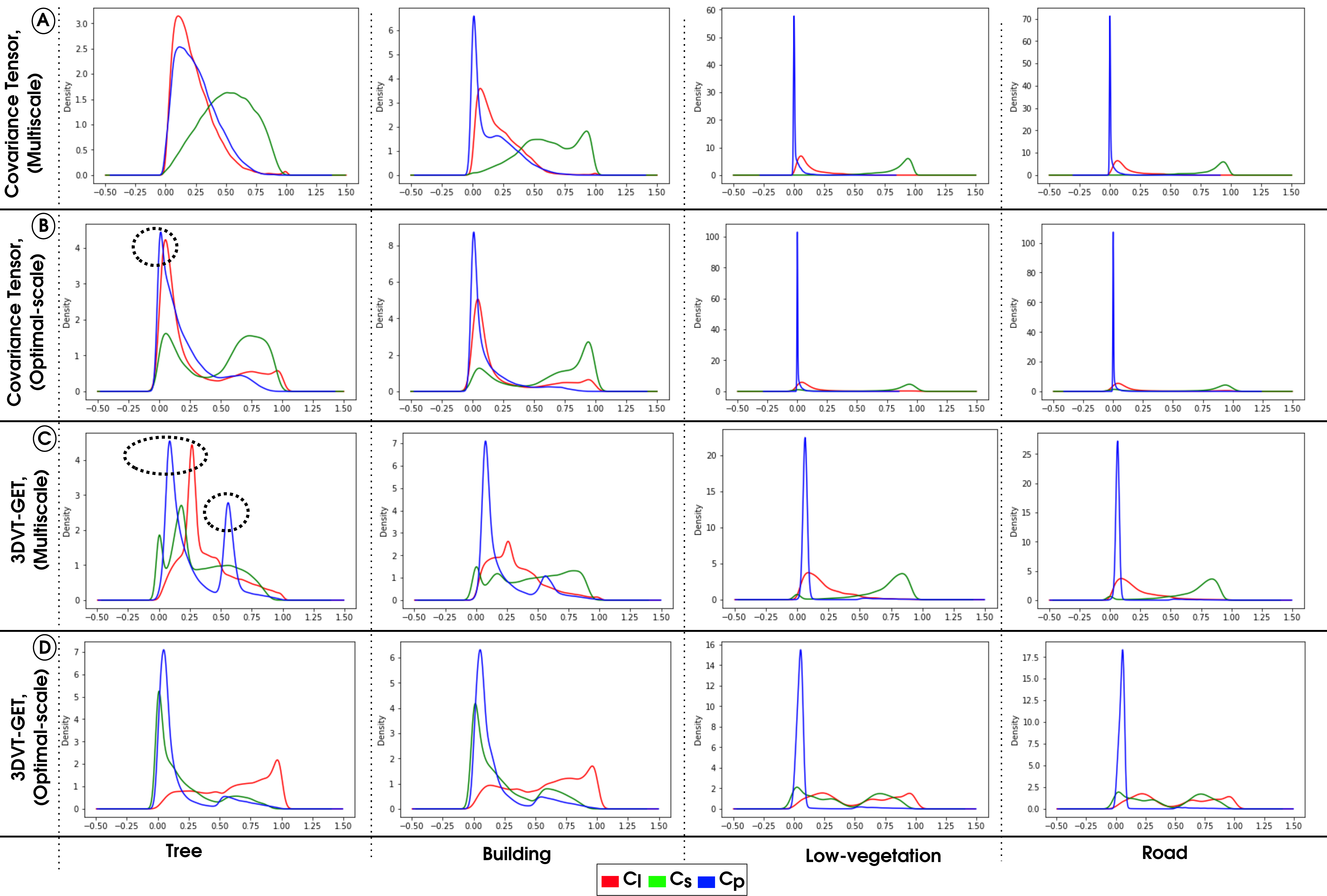}}\\
(i) Area-2\\\vspace{0.1cm}
\fbox{\includegraphics[width=\columnwidth]{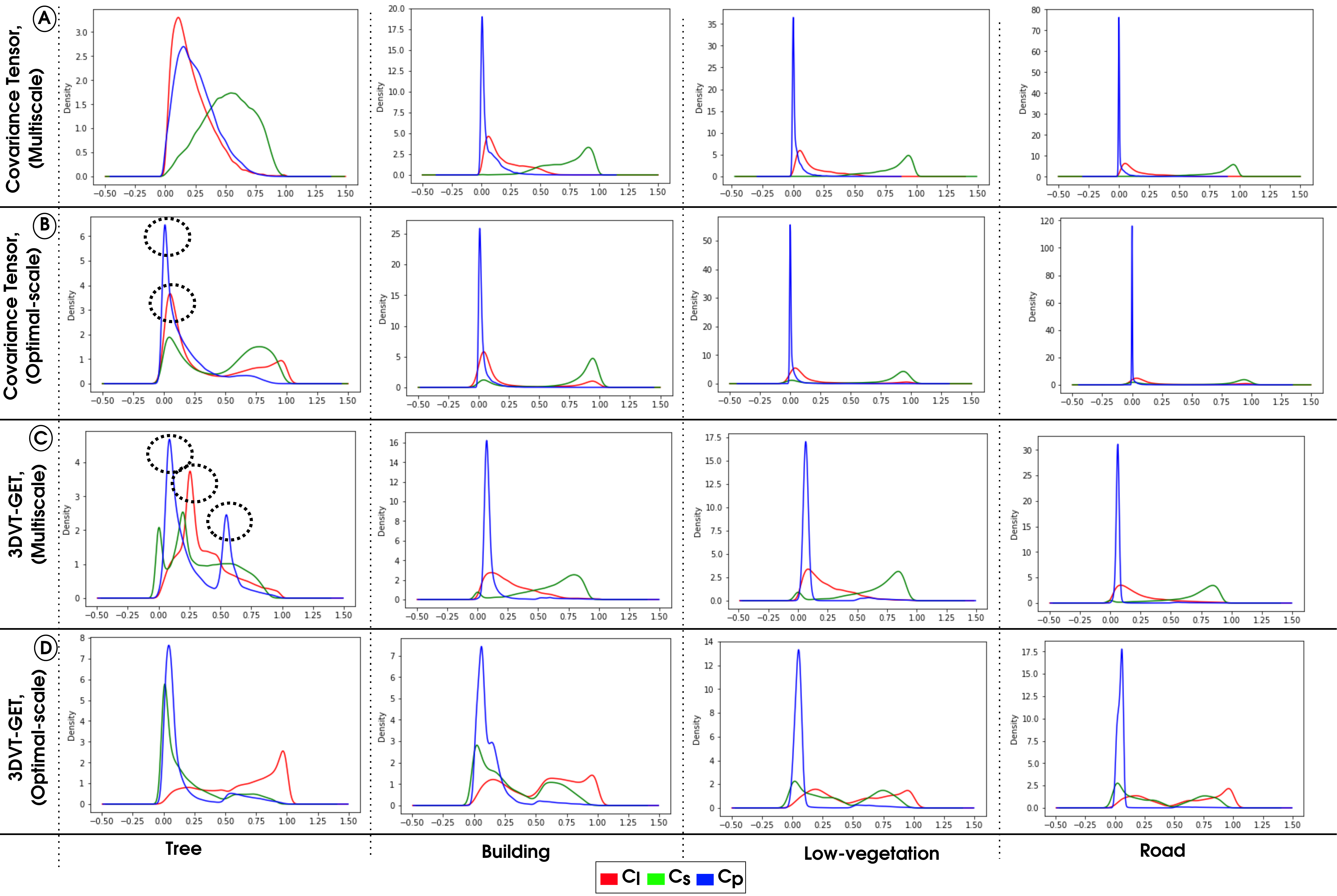}}\\
(ii) Area-3
\caption{\label{fig:a2a3pdf}
Semantic class-wise probability density functions (pdfs) of saliency maps ($\cl$, $\cs$, $\cp$) for (i) Area-2 and (ii) Area-3 of Vaihingen site. The dotted circles show the distinctive behavior of the pdfs in our selected local geometric descriptors with appropriate multiscale approach, for the tree class.}
\end{figure} 

\begin{table}[htbp] 
\scriptsize
\caption{The count distribution of 3D points in each semantic class in point clouds of Vaihingen site from the ISPRS benchmark data~\cite{rottensteiner2012isprs}.}
\label{table:a2a3-sim}
\centering
\begin{tabular}{|c||c|c|c|c|} \hline
  Dataset 
& \textbf{Tree} &  \textbf{Building} & \textbf{Low-vegetation} & \textbf{Road}\\
& (\#points) &(\#points) &(\#points) &(\#points) \\\hline\hline
\multicolumn{1}{|c|} {\textbf{Area-2} with 266,675 points for area of $225m \times 256m$}
& 107,485 & 69,097 & 38,866 & 51,227  \\
 \multicolumn{1}{|c|}{(point sampling density of 2.76 points/$m^2$)}
&&&&\\\hline
 
 \multicolumn{1}{|c|}{\textbf{Area-3} with 323,896 points for area of $193m\times 277m$} 
 & 135,449 & 69,563 & 35,512 & 83,372  \\
 \multicolumn{1}{|c|}{(point sampling density of 2.54 points/$m^2$)}
 &&&&\\\hline

\end{tabular}
\normalsize
\end{table}

\subsection{Visualizations and Comparative Analysis}\label{sec:expt-analysis}
All $\sg$ use rendering of points in the barycentric coordinate space, using fixed position coordinates for the reference triangle in the barycentric coordinate space. For point rendering, the position coordinates of the points in the $\sg$ are computed as the weighted sum of the vertices reference triangle, where the saliency map is used as weights (\S Appendix~\ref{sec:bg-bcs}). Fixing the position coordinates of the reference triangle is important for computing image-based differences between signatures.

\section{Results and Discussion}\label{sec:results}
\begin{figure}[htpb] 
\centering  
\fbox{\includegraphics[width=\columnwidth]{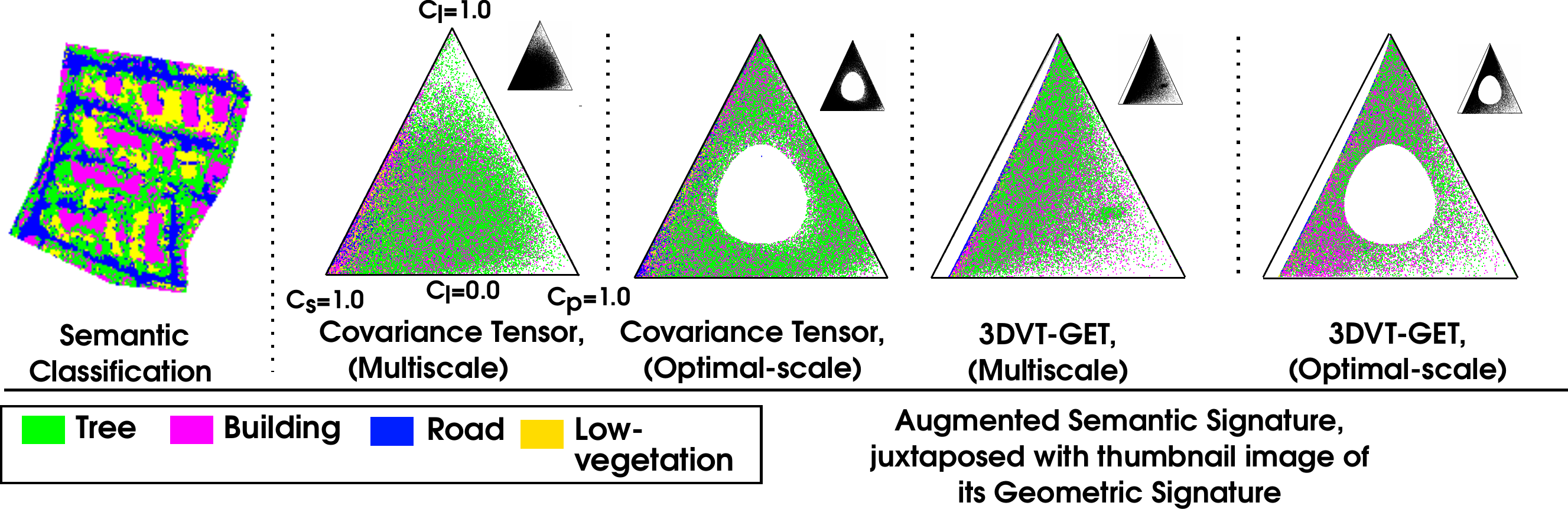}}\\
(i) Area-2 \\\vspace{0.1cm}
\fbox{\includegraphics[width=\columnwidth]{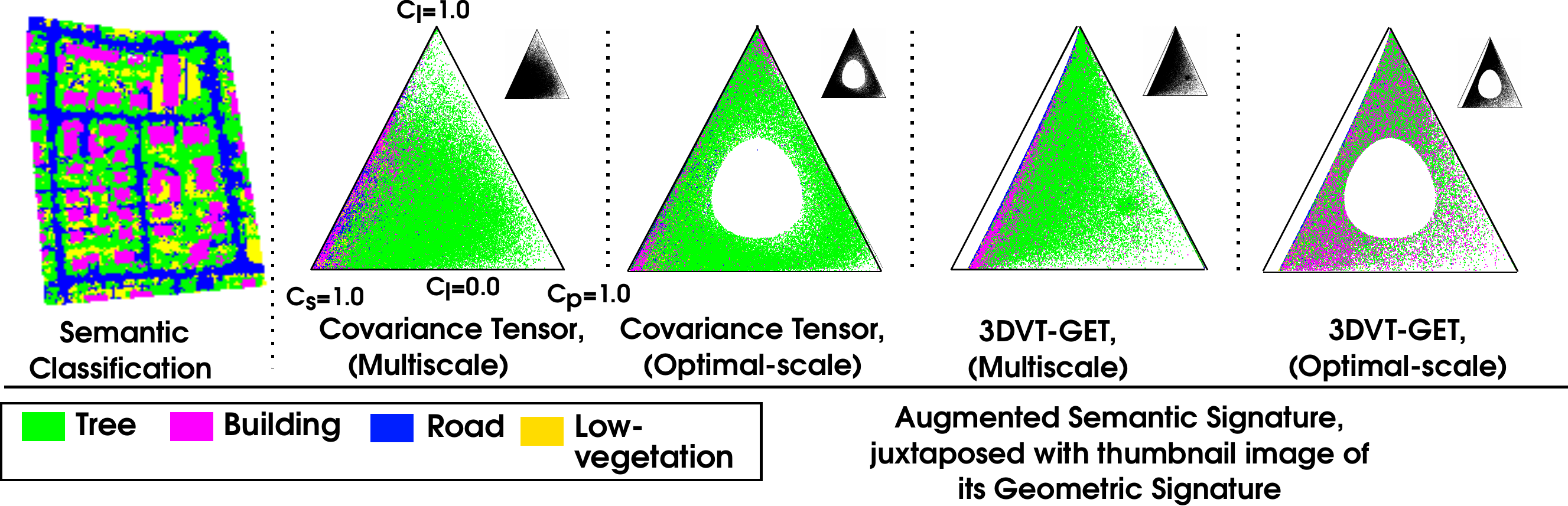}}\\
(ii) Area-3 \\
\caption{\label{fig:a2a3-augsig}
Augmented semantic signature with geometric signature inset, for (i) Area-2 and (ii) Area-3 of Vaihingen site show the augmented semantic signatures using different local geometric descriptors and multiscale approaches. }
\end{figure} 

The pdfs of semantic class-wise saliency maps (\S Figure~\ref{fig:a2a3pdf})  explain the patterns observed in signatures. The pdfs of saliency maps of building and tree classes show multimodal distributions, which manifest as highly \textit{dispersed} points in the signatures. The saliency map using \mulscaleapproach~$\vtget$ gives more (critical) point-type features in the tree class~\cite{sreevalsannair2018contour} and thus the $\sg$ and $\sas$ of the LGD show more points in the vertex corresponding to $\cs=1$ (lower-right corner of the reference triangle, \S Figure~\ref{fig:a2a3-augsig}).

\begin{figure}[htbp]
\centering
\fbox{\includegraphics[width=\columnwidth]{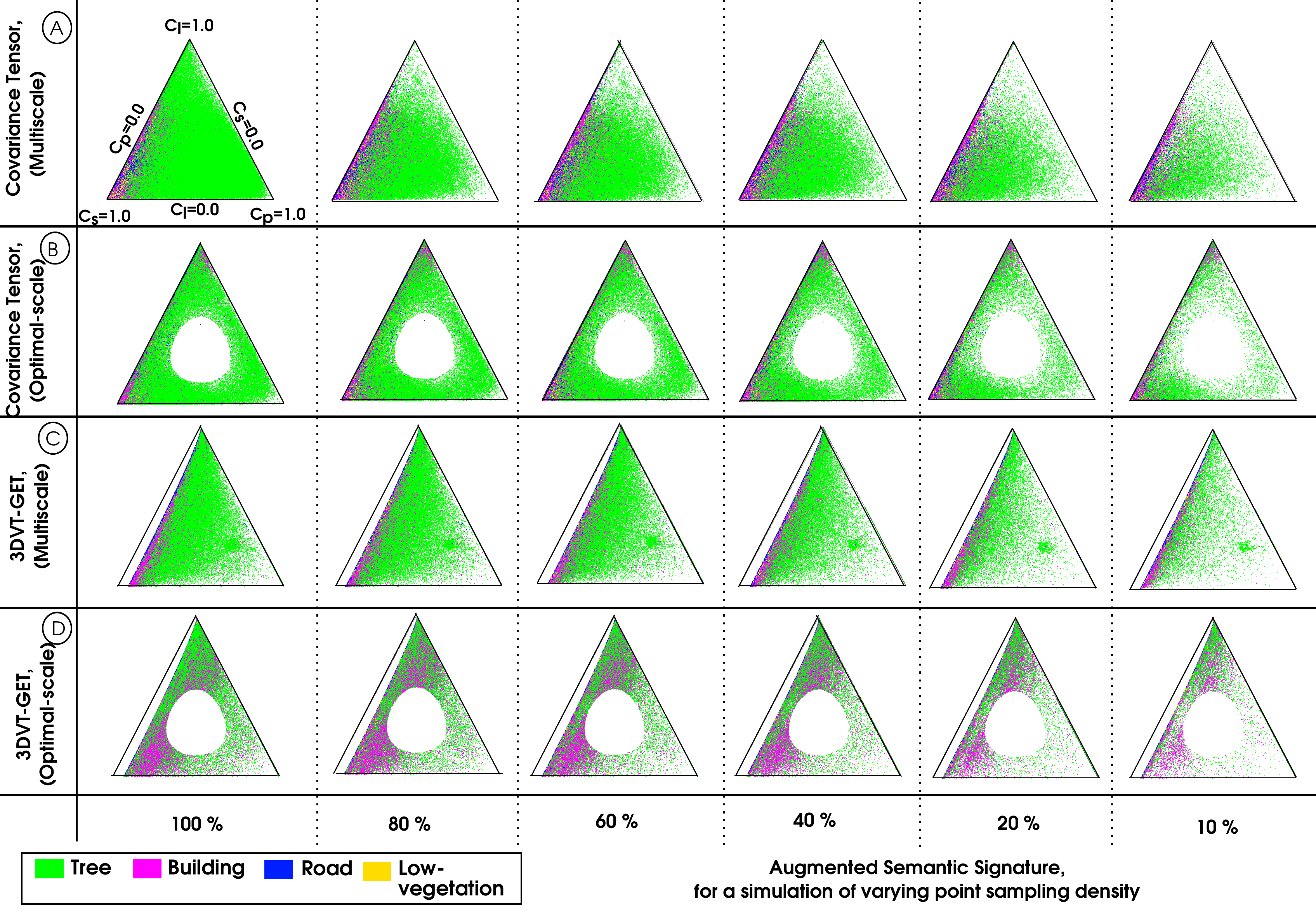}}
\caption{\label{fig:a3sampling}
Effects of sampling density in point cloud in the augmented semantic signatures of Area-3 of the Vaihingen site, using uniform spatial sampling. }
\end{figure}

The observable differences in the signatures are caused by the difference in the uncertainty in geometric classification. The key aspect of our work is to use our proposed signatures with point clouds with different sampling densities. Hence, we study the resilience of our signatures with respect to change in sampling density (\S Figure~\ref{fig:a3sampling}). We observe that the $\sg$ and $\sas$ exhibit sparsification only after the sampling density falls below 60\% of the original point cloud, when using uniform sampling in the 3D cloud, demonstrating the \textit{resilience} of signatures to sampling density reduction. Thus, our proposed signatures can be effectively used for comparing two point clouds with $\lesssim$40\% relative difference in point count. 

\mysubheading{Distance Metrics:}
Our work is generalized for comparing signatures and finding integrated distance between any two point clouds. For instance, $\bdist$ values between Area-3 and Area-2, which is at 82\% point sampling density, are 0.2269, 0.2432, and 0.228 using \mulscaleapproach~$\cov$, \optscaleapproach~$\cov$, and \mulscaleapproach~$\vtget$, respectively. Also, $\kld$, $\hmd$, and the symmetric $\fvd$ (using \mulscaleapproach~$\cov$) between Area-2 and Area-3 are 0.0382, 0.1609, and 0.0835, respectively. Since $\bdist$ is bounded in [0,1], we can say that the two point clouds have relatively small integrated distance. Further work in making our proposed measures bounded or dimensionless will improve the usability of our distance measures in the generalized application.

\mysubheading{Discussions:} 
The state-of-the-art Fisher vector representation is high-dimensional, owing to which the representation cannot have an easy visual representation as our proposed geometric signatures. Overall, the Bhattacharyya distance between signatures $\bdist$ gives an effective integrated measure of change in semantic composition and geometric uncertainty.

However, there are three inherent limitations in our signatures. Firstly, it does not preserve the spatial contextual information which is a disadvantage for spatial inferential tasks. Secondly, the geometric signature depends on the local geometric descriptor. The effectiveness of the signature in demonstrating differences relies on the clear separability of the classes in the pdfs of the saliency maps (\S Figure~\ref{fig:a2a3pdf}). Thirdly, our $\sg$ suffer from information loss owing to the non-injective mapping of Cartesian coordinate space to barycentric coordinate space. This causes overlap of points in barycentric space, and hence, occlusion of points in the visualization of the signature. Also, the occlusion of signatures is determined by the order of point rendering, which indirectly affects $\sas$. However, this shortcoming can be alleviated by using appropriate colormaps, which exploit transparencies for $\sas$ to reduce occlusion.

\section{Conclusions}
Three-dimensional point clouds acquired using LiDAR technology are useful for topographical study, given the speed and accuracy of data acquisition LiDAR can provide. By the same token, LiDAR point clouds also come with challenges in its analysis given high unpredictability and variability in the type of objects encountered in outdoor environments, especially. In this work, we have proposed a geometric signature $\sg$ of an airborne LiDAR point cloud, which can be visualized and used for comparing datasets. $\sg$ encapsulates the probabilistic geometric classification of the point cloud into line-, surface-, and (critical) point-type features. Augmented semantic signatures $\sas$ are a variant of $\sg$ that use color corresponding to the semantic class. $\sg$ is generated in image format after converting 3D points to 2D barycentric coordinate representation using the saliency map from the probabilistic geometric classification.

Our proposed signatures are global descriptors of the point cloud, and we have demonstrated its usability in comparing point clouds of two regions in the same (Vaihingen) site. We propose integrated measures of differences in both semantic composition and geometric uncertainty. We have proposed an entropy measure based on probabilistic geometric classification, namely, the saliency map-based Shannon entropy, $\egeom$. We have found that our signatures are resilient for relative differences in sample size to be $\lesssim$40\%.

Overall, we have shown that our novel augmented semantic signatures are effective global descriptors of point clouds for measuring differences in geometric uncertainty and semantic composition between two datasets. This work can be effectively used as generalized integrated measures of distances between any two point clouds.

\section*{Acknowledgments}
The authors are grateful to Satendra Singh for the code refactor that preceded the proposed work and Intel India Research Fellowship. This work has been supported by the Early Career Research Award of the first author, by the Science and Engineering Research Board (SERB), Government of India. This work has used third-party software libraries,  namely, PCL~\cite{rusu20113d}, CGAL~\cite{fabri2009cgal}, OpenCV~\cite{opencv_library}, and Python libraries~\cite{jones2001scipy} for implementations of point cloud processing, geometric processing, and image processing algorithms, respectively. The Vaihingen data set has been published by the German Society for Photogrammetry, Remote Sensing and Geoinformation (DGPF)~\cite{cramer2010dgpf}\footnote{Vaihingen dataset acquisition project: \url{http://www.ifp.uni-stuttgart.de/dgpf/DKEP-Allg.html}}

\appendix
\section{Computations in Barycentric Coordinate System}\label{sec:bg-bcs}
The barycentric coordinate system is an alternative to an affine or cartesian coordinate system, where there is a notion of a \textit{barycenter} or a center of mass. It is a coordinate system described on a simplex, such as a triangle, tetrahedron, etc. In the barycentric coordinate system, the simplex corresponds to its dimensionality $d$ of the cartesian coordinate system, which means, triangle corresponds to a 2D space, tetrahedron corresponds to a 3D space, and so on.

There are four key features of the barycentric coordinate space. Firstly, the representation of a point is a $(d+1)$-tuple for a $d$-dimensional space, unlike a $d$-tuple in an equivalent cartesian coordinate system. That said, in both cases, only $d$ dimensions can be strictly linearly independent, which means in the barycentric coordinate system, there is one dependent dimension. As an example, in a 2D barycentric coordinate, a point $A$ is described as $(x_0(A),x_1(A),x_2(A))$, with $\sum\limits_{i=0}^2 x_i = 1.0$.

\noindent Thus, the second feature is that the coordinates of a point in a barycentric coordinate system have the property of \textit{partition to unity}. The position coordinates of $A$ in the barycentric coordinate system can be rewritten as $(x_0(A),x_1(A),1-(x_0(A)+x_1(A)))$, and similarly, any of the three dimensions can be chosen to be linearly dependent one. Thirdly, for a point in the interior of the simplex, i.e., inside the triangle in 2D case, all coordinates of all points in the simplex lie in the interval $[0,1]$. The fourth feature of barycentric coordinate system is that each coordinate $x_i$ corresponds to a vertex $V_i$ of the $d$-dimensional simplex, where $0\le i \le (d-1)$ and $i\in\mathbb{Z}^+$. Summing up all four features, any property $q(A)$ at a point $A$ with barycentric coordinates $(x_0,x_1,\ldots,x_{d-1})$, will be a \textit{convex combination} of the property at the vertices of the simplex, e.g. $q$ can be the 3D coordinates of $A$, the color at $A$, etc. Thus, \\\centerline{$q(A)=\sum\limits_{i=0}^{d-1} x_i.q(V_i)$}.

We can see that the property of partition to unity is analogous to the probabilities in the saliency map. Owing to the analogy of barycentric coordinates with the sum of likelihoods in probabilistic geometric classification, i.e., $\cl(A)+\cs(A)+\cp(A)=1.0$, at point A, we can map the saliency values as barycentric coordinates. Thus, we propose the transformation of point cloud from 3D cartesian coordinate system to 2D barycentric space to generate the geometric signature of the point cloud. 

\section{Multiscale Computations for Local Geometric Descriptors}\label{sec:bg-lgd}
Here, we explore the use of multiple scales for both averaging saliency maps across scales, as well as computation of saliency map using an optimal scale based on a novel variant of Shannon entropy using saliency maps themselves (\S Section~\ref{sec:method_signature}). 

\mysubheading{Multiscale Aggregation of Saliency Maps:} We compute saliency maps for line-, surface-, and (critical) point-type features in a point cloud using local geometric descriptors (LGDs). The saliency map is computed as a 3-tuple, using the likelihood of a point being a line-type $\cl$, surface-type $\cs$, and point-type $\cp$ feature. Since these three are the only possibilities of geometric classes in 3D space, we get $\cl+\cs+\cp=1.0$ at each point. The saliency map at each point $x$ is computed using the eigenvalue decomposition of its LGD~\cite{sreevalsannair2017local}, for each scale $\scale$ at each point $x$, as given below, where the eigenvalues of the LGD are sorted as $\lambda_0\ge\lambda_1\ge\lambda_2$:
\begin{eqnarray}\label{eqn:saliency}
  \cl(x,\scale)&=&\eigsum.(\lambda_0(x,\scale)-\lambda_1(x,\scale)), \nonumber\\
  \cs(x,\scale)&=&\eigsum.(2(\lambda_1(x,\scale)-\lambda_2(x,\scale)), \text{  and  } \nonumber\\
  \cp(x,\scale)&=&\eigsum.(3(\lambda_2(x,\scale))), \text{   where   } \nonumber\\
  \eigsum&=&\frac{1}{\lambda_0(x,\scale)+\lambda_1(x,\scale)+\lambda_2(x,\scale)}.
\end{eqnarray}
There are two strategies for aggregating the saliency maps across scales. The saliency maps are averaged across scales, as they are likelihoods of points belonging to a geometric class, and the different scales are mutually exclusive events ~\cite{sreevalsannair2017local}, in which case, for $N_\scale$ scales in $[\scale_{min},\scale_{max}]$\\\centerline{$\salmap(x)=\frac{1}{N_\scale}\sum\limits_{\scale=\scale_{min}}^{\scale=\scale_{max}}\{\cl(x,\scale),\cs(x,\scale),\cp(x,\scale)\}$.} The second strategy is to use optimal scale, which is determined based on minimum value of chosen Shannon entropy. In the case of optimal scale, \\\centerline{$\salmap(x)=\{\cl(x,\optscale),\cs(x,\optscale),\cp(x,\optscale)\}$.}

\bibliographystyle{unsrt}
\bibliography{papers_consolidated}

\begin{thebibliography}{10}

\bibitem{rottensteiner2009status}
Franz Rottensteiner.
\newblock {Status and further prospects of object extraction from image and
  laser data}.
\newblock In {\em 2009 Joint Urban Remote Sensing Event}, pages 1--10. IEEE,
  2009.

\bibitem{jalobeanu2014uncertainty}
Andr{\'e} Jalobeanu, Angela~M Kim, Scott~C Runyon, RC~Olsen, and Fred~A Kruse.
\newblock {Uncertainty assessment and probabilistic change detection using
  terrestrial and airborne LiDAR}.
\newblock In {\em Laser Radar Technology and Applications XIX; and Atmospheric
  Propagation XI}, volume 9080, page 90800S. International Society for Optics
  and Photonics, 2014.

\bibitem{ben20183dmfv}
Yizhak Ben-Shabat, Michael Lindenbaum, and Anath Fischer.
\newblock {3dmfv: Three-dimensional point cloud classification in real-time
  using convolutional neural networks}.
\newblock {\em IEEE Robotics and Automation Letters}, 3(4):3145--3152, 2018.

\bibitem{rottensteiner2012isprs}
Franz Rottensteiner, Gunho Sohn, Jaewook Jung, M~Gerke, Caroline Baillard,
  Sebastien Benitez, and Uwe Breitkopf.
\newblock {The ISPRS benchmark on urban object classification and 3D building
  reconstruction}.
\newblock {\em ISPRS Annals of Photogrammetry, Remote Sensing and Spatial
  Information Sciences I-3}, pages 293--298, 2012.

\bibitem{sreevalsannair2017local}
Jaya Sreevalsan-Nair and Beena Kumari.
\newblock {\em {Local Geometric Descriptors for Multi-Scale Probabilistic Point
  Classification of Airborne LiDAR Point Clouds}}, pages 175--200.
\newblock Springer Cham, Mathematics and Visualization, 2017.

\bibitem{kumari2015interactive}
Beena Kumari and Jaya Sreevalsan-Nair.
\newblock {An interactive visual analytic tool for semantic classification of
  3D urban LiDAR point cloud}.
\newblock In {\em Proceedings of the 23rd SIGSPATIAL International Conference
  on Advances in Geographic Information Systems}, page~73. ACM, 2015.

\bibitem{weinmann2015semantic}
Martin Weinmann, Boris Jutzi, Stefan Hinz, and Cl\'ement Mallet.
\newblock {Semantic point cloud interpretation based on optimal neighborhoods,
  relevant features and efficient classifiers}.
\newblock {\em {ISPRS Journal of Photogrammetry and Remote Sensing}},
  105:286--304, 2015.

\bibitem{rusu2008aligning}
Radu~Bogdan Rusu, Nico Blodow, Zoltan~Csaba Marton, and Michael Beetz.
\newblock {Aligning point cloud views using persistent feature histograms}.
\newblock In {\em 2008 IEEE/RSJ International Conference on Intelligent Robots
  and Systems}, pages 3384--3391. IEEE, 2008.

\bibitem{mahmoudi2009three}
Mona Mahmoudi and Guillermo Sapiro.
\newblock {Three-dimensional point cloud recognition via distributions of
  geometric distances}.
\newblock {\em Graphical Models}, 71(1):22--31, 2009.

\bibitem{velizhev2012implicit}
Alexander Velizhev, Roman Shapovalov, and Konrad Schindler.
\newblock {Implicit shape models for object detection in 3D point clouds}.
\newblock In {\em International Society of Photogrammetry and Remote Sensing
  Congress}, volume~2, 2012.

\bibitem{brodu20123d}
Nicolas Brodu and Dimitri Lague.
\newblock {3D terrestrial lidar data classification of complex natural scenes
  using a multi-scale dimensionality criterion: Applications in geomorphology}.
\newblock {\em ISPRS Journal of Photogrammetry and Remote Sensing},
  68:121--134, 2012.

\bibitem{gross2006extraction}
Hermann Gross and Ulrich Thoennessen.
\newblock {Extraction of lines from laser point clouds}.
\newblock In {\em Symposium of ISPRS Commission III: Photogrammetric Computer
  Vision PCV06. International Archives of Photogrammetry, Remote Sensing and
  Spatial Information Sciences}, volume~36, pages 86--91, 2006.

\bibitem{hoppe1992surface}
Hugues Hoppe, Tony DeRose, Tom Duchamp, John McDonald, and Werner Stuetzle.
\newblock {Surface Reconstruction from Unorganized Points}.
\newblock {\em SIGGRAPH Comput. Graph.}, 26(2):71--78, July 1992.

\bibitem{sreevalsannair2018contour}
Jaya Sreevalsan-Nair, Akshay Jindal, and Beena Kumari.
\newblock {Contour extraction in buildings in airborne lidar point clouds using
  multiscale local geometric descriptors and visual analytics}.
\newblock {\em IEEE Journal of Selected Topics in Applied Earth Observations
  and Remote Sensing}, 11(7):2320--2335, 2018.

\bibitem{sun2009concise}
Jian Sun, Maks Ovsjanikov, and Leonidas Guibas.
\newblock {A concise and provably informative multi-scale signature based on
  heat diffusion}.
\newblock In {\em Computer Graphics Forum}, volume~28, pages 1383--1392. Wiley
  Online Library, 2009.

\bibitem{li2005multiscale}
Xinju Li and Igor Guskov.
\newblock {Multiscale Features for Approximate Alignment of Point-based
  Surfaces}.
\newblock In {\em Symposium on geometry processing}, volume 255, page 217.
  Citeseer, 2005.

\bibitem{shen2018mining}
Yiru Shen, Chen Feng, Yaoqing Yang, and Dong Tian.
\newblock {Mining point cloud local structures by kernel correlation and graph
  pooling}.
\newblock In {\em Proceedings of the IEEE conference on computer vision and
  pattern recognition}, pages 4548--4557, 2018.

\bibitem{guo2013rotational}
Yulan Guo, Ferdous Sohel, Mohammed Bennamoun, Min Lu, and Jianwei Wan.
\newblock {Rotational projection statistics for 3D local surface description
  and object recognition}.
\newblock {\em International journal of computer vision}, 105(1):63--86, 2013.

\bibitem{guo2015novel}
Yulan Guo, Ferdous Sohel, Mohammed Bennamoun, Jianwei Wan, and Min Lu.
\newblock {A novel local surface feature for 3D object recognition under
  clutter and occlusion}.
\newblock {\em Information Sciences}, 293:196--213, 2015.

\bibitem{dong2017novel}
Zhen Dong, Bisheng Yang, Yuan Liu, Fuxun Liang, Bijun Li, and Yufu Zang.
\newblock {A novel binary shape context for 3D local surface description}.
\newblock {\em ISPRS Journal of Photogrammetry and Remote Sensing},
  130:431--452, 2017.

\bibitem{perronnin2010improving}
Florent Perronnin, Jorge S{\'a}nchez, and Thomas Mensink.
\newblock {Improving the fisher kernel for large-scale image classification}.
\newblock In {\em European conference on computer vision}, pages 143--156.
  Springer, 2010.

\bibitem{jegou2010aggregating}
Herv{\'e} J{\'e}gou, Matthijs Douze, Cordelia Schmid, and Patrick P{\'e}rez.
\newblock {Aggregating local descriptors into a compact image representation}.
\newblock In {\em CVPR 2010-23rd IEEE Conference on Computer Vision \& Pattern
  Recognition}, pages 3304--3311. IEEE Computer Society, 2010.

\bibitem{jaakkola1999exploiting}
Tommi Jaakkola and David Haussler.
\newblock {Exploiting generative models in discriminative classifiers}.
\newblock In {\em Advances in neural information processing systems}, pages
  487--493, 1999.

\bibitem{sanchez2013image}
Jorge S{\'a}nchez, Florent Perronnin, Thomas Mensink, and Jakob Verbeek.
\newblock {Image classification with the fisher vector: Theory and practice}.
\newblock {\em International journal of computer vision}, 105(3):222--245,
  2013.

\bibitem{angelina2018pointnetvlad}
Mikaela Angelina~Uy and Gim Hee~Lee.
\newblock {PointNetVLAD: Deep point cloud based retrieval for large-scale place
  recognition}.
\newblock In {\em Proceedings of the IEEE Conference on Computer Vision and
  Pattern Recognition}, pages 4470--4479, 2018.

\bibitem{cignoni1998metro}
Paolo Cignoni, Claudio Rocchini, and Roberto Scopigno.
\newblock {Metro: measuring error on simplified surfaces}.
\newblock In {\em Computer Graphics Forum}, volume~17, pages 167--174. Wiley
  Online Library, 1998.

\bibitem{aspert2002mesh}
Nicolas Aspert, Diego Santa-Cruz, and Touradj Ebrahimi.
\newblock {Mesh: Measuring errors between surfaces using the hausdorff
  distance}.
\newblock In {\em Proceedings. IEEE international conference on multimedia and
  expo}, volume~1, pages 705--708. IEEE, 2002.

\bibitem{richter2013out}
Rico Richter, Jan~Eric Kyprianidis, and J{\"u}rgen D{\"o}llner.
\newblock {Out-of-Core GPU-based Change Detection in Massive 3D Point Clouds}.
\newblock {\em Transactions in GIS}, 17(5):724--741, 2013.

\bibitem{memoli2004comparing}
Facundo M{\'e}moli and Guillermo Sapiro.
\newblock {Comparing point clouds}.
\newblock In {\em Proceedings of the 2004 Eurographics/ACM SIGGRAPH symposium
  on Geometry processing}, pages 32--40. ACM, 2004.

\bibitem{tran2018integrated}
Thi Tran, Camillo Ressl, and Norbert Pfeifer.
\newblock {Integrated change detection and classification in urban areas based
  on airborne laser scanning point clouds}.
\newblock {\em Sensors}, 18(2):448, 2018.

\bibitem{demantke2011dimensionality}
J{\'e}r{\^o}me Demantk{\'e}, Cl{\'e}ment Mallet, Nicolas David, and Bruno
  Vallet.
\newblock {Dimensionality based Scale Selection in 3D LiDAR Point Clouds}.
\newblock {\em The International Archives of the Photogrammetry, Remote Sensing
  and Spatial Information Sciences}, 38(Part 5):W12, 2011.

\bibitem{keller2011extracting}
Patric Keller, Oliver Kreylos, Marek Vanco, Martin Hering-Bertram, Eric~S
  Cowgill, Louise~H Kellogg, Bernd Hamann, and Hans Hagen.
\newblock {Extracting and visualizing structural features in environmental
  point cloud LiDaR data sets}.
\newblock In {\em {Topological Methods in Data Analysis and Visualization}},
  pages 179--192. Springer, 2011.

\bibitem{sreevalsannair2017using}
Jaya Sreevalsan-Nair and Akshay Jindal.
\newblock {Using gradients and tensor voting in 3D local geometric descriptors
  for feature detection in airborne lidar point clouds in urban regions}.
\newblock In {\em 2017 IEEE International Geoscience and Remote Sensing
  Symposium (IGARSS)}, pages 5881--5884. IEEE, 2017.

\bibitem{wang2014multiscale}
Zhen Wang, Liqiang Zhang, Tian Fang, P~Takis Mathiopoulos, Xiaohua Tong, Huamin
  Qu, Zhiqiang Xiao, Fang Li, and Dong Chen.
\newblock {A multiscale and hierarchical feature extraction method for
  terrestrial laser scanning point cloud classification}.
\newblock {\em IEEE Transactions on Geoscience and Remote Sensing},
  53(5):2409--2425, 2014.

\bibitem{zhang2016multilevel}
Zhenxin Zhang, Liqiang Zhang, Xiaohua Tong, P~Takis Mathiopoulos, Bo~Guo,
  Xianfeng Huang, Zhen Wang, and Yuebin Wang.
\newblock {A multilevel point-cluster-based discriminative feature for ALS
  point cloud classification}.
\newblock {\em IEEE Transactions on Geoscience and Remote Sensing},
  54(6):3309--3321, 2016.

\bibitem{niemeyer2014contextual}
Joachim Niemeyer, Franz Rottensteiner, and Uwe Soergel.
\newblock {Contextual classification of lidar data and building object
  detection in urban areas}.
\newblock {\em ISPRS journal of photogrammetry and remote sensing},
  87:152--165, 2014.

\bibitem{hackel2016fast}
Timo Hackel, Jan~D Wegner, and Konrad Schindler.
\newblock {Fast semantic segmentation of 3D point clouds with strongly varying
  density}.
\newblock {\em ISPRS annals of the photogrammetry, remote sensing and spatial
  information sciences}, 3(3):177--184, 2016.

\bibitem{thomas2018semantic}
Hugues Thomas, Fran{\c{c}}ois Goulette, Jean-Emmanuel Deschaud, and Beatriz
  Marcotegui.
\newblock {Semantic classification of 3d point clouds with multiscale spherical
  neighborhoods}.
\newblock In {\em 2018 International Conference on 3D Vision (3DV)}, pages
  390--398. IEEE, 2018.

\bibitem{battke1997fast}
Henrik Battke, Detlev Stalling, and Hans-Christian Hege.
\newblock {Fast line integral convolution for arbitrary surfaces in 3D}.
\newblock In {\em Visualization and Mathematics}, pages 181--195. Springer,
  1997.

\bibitem{digne2011scale}
Julie Digne, Jean-Michel Morel, Charyar-Mehdi Souzani, and Claire Lartigue.
\newblock {Scale space meshing of raw data point sets}.
\newblock In {\em Computer Graphics Forum}, volume~30, pages 1630--1642. Wiley
  Online Library, 2011.

\bibitem{zalama2011effective}
Eduardo Zalama, Jaime G{\'o}mez-Garc{\'\i}a-Bermejo, Jos{\'e} Llamas, and
  Roberto Medina.
\newblock {An effective texture mapping approach for 3D models obtained from
  laser scanner data to building documentation}.
\newblock {\em Computer-Aided Civil and Infrastructure Engineering},
  26(5):381--392, 2011.

\bibitem{langer2008generalized}
Torsten Langer et~al.
\newblock {\em {On Generalized Barycentric Coordinates and Their Applications
  in Geometric Modeling}}.
\newblock PhD thesis, Universit{\"a}t des Saarlandes Saarbr{\"u}cken, 2008.

\bibitem{medioni2000tensor}
G{\'e}rard Medioni, Chi-Keung Tang, and Mi-Suen Lee.
\newblock {Tensor voting: Theory and applications}.
\newblock {\em Proceedings of RFIA, Paris, France}, 3, 2000.

\bibitem{park2012multi}
Min~Ki Park, Seung~Joo Lee, and Kwan~H Lee.
\newblock {Multi-scale Tensor Voting for Feature Extraction from Unstructured
  Point Clouds}.
\newblock {\em Graphical Models}, 74(4):197--208, 2012.

\bibitem{wang2013anisotropic}
Shengfa Wang, Tingbo Hou, Shuai Li, Zhixun Su, and Hong Qin.
\newblock {Anisotropic Elliptic PDEs for Feature Classification}.
\newblock {\em Visualization and Computer Graphics, IEEE Transactions on},
  19(10):1606--1618, 2013.

\bibitem{tong2004first}
Wai-Shun Tong, Chi-Keung Tang, Philippos Mordohai, and G{\'e}rard Medioni.
\newblock {First order augmentation to tensor voting for boundary inference and
  multiscale analysis in 3D}.
\newblock {\em IEEE transactions on pattern analysis and machine intelligence},
  26(5):594--611, 2004.

\bibitem{weinmann2014semantic}
Martin Weinmann, Boris Jutzi, and Cl{\'e}ment Mallet.
\newblock {Semantic 3D scene interpretation: A framework combining optimal
  neighborhood size selection with relevant features}.
\newblock {\em ISPRS Annals of the Photogrammetry, Remote Sensing and Spatial
  Information Sciences}, 2(3):181, 2014.

\bibitem{qianwen2014comparison}
{Qianwen Zhang and Roxanne L. Canosa}.
\newblock {A comparison of histogram distance metrics for content-based image
  retrieval}.
\newblock In Qian Lin, Jan~Philip Allebach, and Zhigang Fan, editors, {\em
  Imaging and Multimedia Analytics in a Web and Mobile World 2014}, volume
  9027, pages 154 -- 162. International Society for Optics and Photonics, SPIE,
  2014.

\bibitem{naik2009scale}
Nikhil Naik, Sanmay Patil, and Madhuri Joshi.
\newblock {A scale adaptive tracker using hybrid color histogram matching
  scheme}.
\newblock In {\em 2009 Second International Conference on Emerging Trends in
  Engineering \& Technology}, pages 279--284. IEEE, 2009.

\bibitem{rusu20113d}
Radu~Bogdan Rusu and Steve Cousins.
\newblock {3D is here: Point Cloud Library (PCL)}.
\newblock In {\em {IEEE International Conference on Robotics and Automation
  (ICRA)}}, Shanghai, China, May 9-13 2011.

\bibitem{fabri2009cgal}
Andreas Fabri and Sylvain Pion.
\newblock {CGAL: The computational geometry algorithms library}.
\newblock In {\em Proceedings of the 17th ACM SIGSPATIAL international
  conference on advances in geographic information systems}, pages 538--539,
  2009.

\bibitem{opencv_library}
G.~Bradski.
\newblock {OpenCV 2.4.11.0 Documentation}.
\newblock {\em {Dr. Dobb's Journal of Software Tools}}, 2000.

\bibitem{jones2001scipy}
Eric Jones, Travis Oliphant, Pearu Peterson, et~al.
\newblock {SciPy: Open source scientific tools for Python}.
\newblock 2001.

\bibitem{cramer2010dgpf}
Michael Cramer.
\newblock {The DGPF-test on digital airborne camera evaluation--overview and
  test design}.
\newblock {\em Photogrammetrie-Fernerkundung-Geoinformation}, 2010(2):73--82,
  2010.

\end{thebibliography}
\end{document}